\documentclass[10pt,twocolumn,letterpaper]{article}
\usepackage[pagenumbers]{cvpr}  
\def\confName{CVPR}
\def\confYear{2025}

\usepackage{amsmath, amssymb, amscd, amsthm, amsfonts}
\usepackage{bm}

\usepackage{graphicx} % Required for inserting images
\usepackage{algorithm}
\usepackage{algorithmic}
\usepackage{booktabs}
\usepackage{multirow}
\usepackage{amsmath,epstopdf,amssymb,color,url,multirow,multicol,verbatim,threeparttable,diagbox,makecell,booktabs,color,colortbl}
\definecolor{light-gray}{gray}{0.82}
\definecolor{aliceblue}{rgb}{0.94,0.97,1.0}

\definecolor{cvprgreen}{rgb}{0.10, 0.52, 0.27}

\definecolor{cvprblue}{rgb}{0.21,0.49,0.74}
\definecolor{darkblue}{RGB}{25,25,180} % dark blue color
\definecolor{darkred}{RGB}{180,0,0} % dark red color
\definecolor{darkgreen}{RGB}{0,180,0} % dark red color
\definecolor{midblue}{RGB}{25,25,112} % mid blue color
\definecolor{darkyellow}{RGB}{240,180,0} % dark yellow color
\usepackage{hyperref}       % hyperlinks
\hypersetup{
colorlinks=true,
linkcolor=darkred,
filecolor=darkred,      
citecolor=cvprblue,
}

\title{Generalizing Deepfake Video Detection with Plug-and-Play: Video-Level Blending and Spatiotemporal Adapter Tuning}

\author{Zhiyuan Yan$^{1,2}$, Yandan Zhao$^2$, Shen Chen$^2$, Mingyi Guo$^1$, Xinghe Fu$^2$,\\ Taiping Yao$^2$ Shouhong Ding$^2$, Li Yuan$^{1\dagger}$ \\
        School of Electronic and Computer Engineering, Peking University$^1$,\\
        Tencent Youtu Lab$^2$,
}

\begin{document}

\maketitle

% recommend 70--150 words
% In recent years, the generalizability of deepfake detectors has become a major challenge for applications. 
\begin{abstract}
Three key challenges hinder the development of current deepfake video detection:
(1) Temporal features can be complex and diverse: how can we identify general temporal artifacts to enhance model generalization?
(2) Spatiotemporal models often lean heavily on one type of artifact and ignore the other: how can we ensure balanced learning from both?
(3) Videos are naturally resource-intensive: how can we tackle efficiency without compromising accuracy?
This paper attempts to tackle the three challenges jointly.
\textbf{First}, inspired by the notable generality of using image-level blending data for image forgery detection, we investigate whether and how video-level blending can be effective in video. 
We then perform a thorough analysis and identify a previously underexplored temporal forgery artifact: \underline{F}acial \underline{F}eature \underline{D}rift (FFD), which commonly exists across different forgeries.
To reproduce FFD, we then propose a novel \textbf{V}ideo-level \textbf{B}lending data (\textbf{VB}), which is implemented by blending the original image and its warped version frame-by-frame, serving as a hard negative sample to mine more general artifacts. 
\textbf{Second}, we carefully design a lightweight \textbf{S}patio\textbf{t}emporal \textbf{A}dapter (\textbf{StA}) to equip a pretrained image model with the ability to capture both spatial and temporal features jointly and efficiently.
StA is designed with two-stream 3D-Conv with varying kernel sizes, allowing it to process spatial and temporal features separately.
% Both our VB and StA can be applied plug-and-play. 
Extensive experiments validate the effectiveness of the proposed methods; and show our approach can generalize well to previously unseen forgery videos, even the latest generation methods.

\end{abstract}

\begin{figure}[!t] %H为当前位置，!htb为忽略美学标准，htbp为浮动图形
\centering %图片居中
\includegraphics[width=0.47\textwidth]{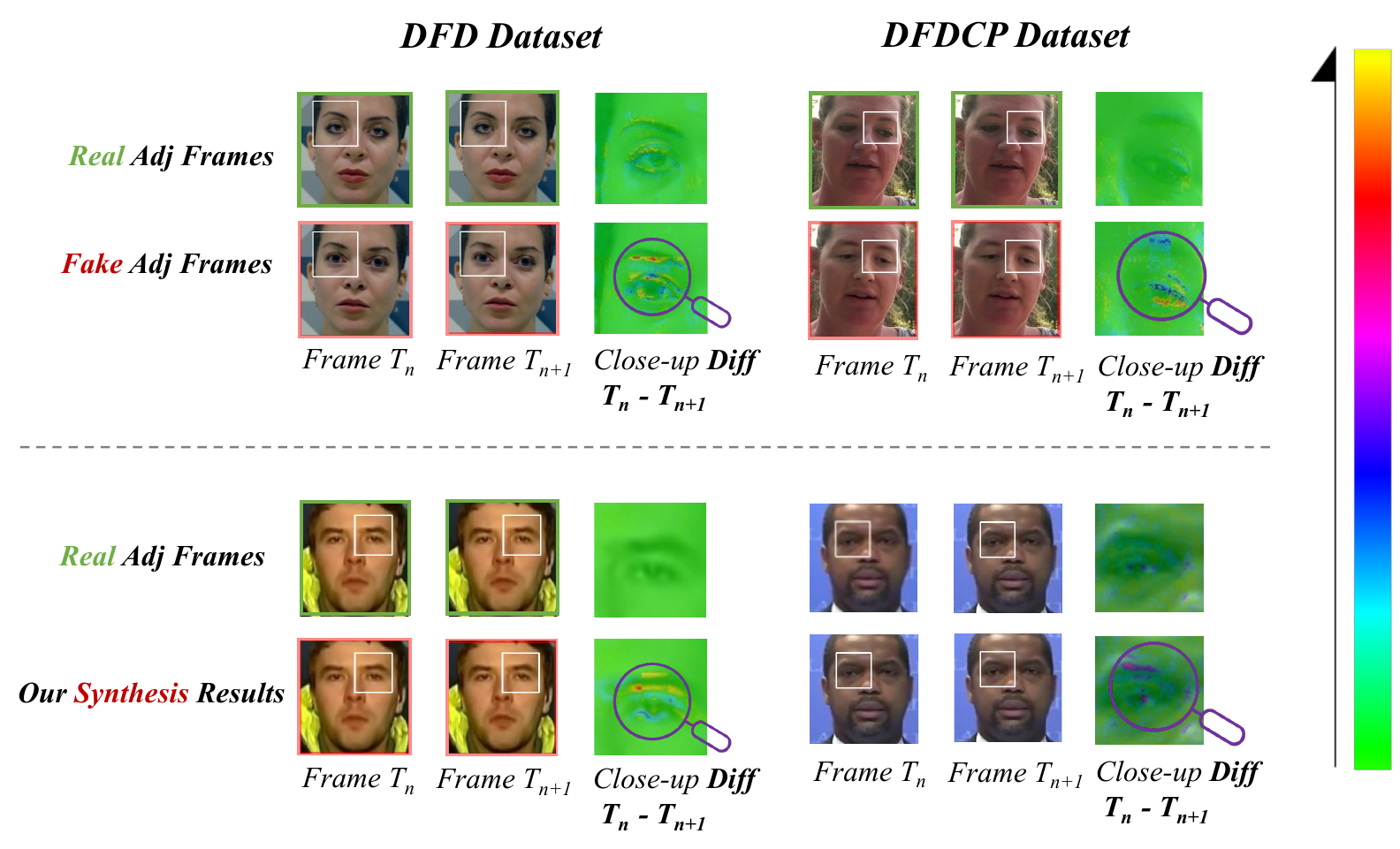} 
\caption{
Illustration of Facial Feature Drift (FFD) phenomena in deepfake videos.
We empirically select two relatively static consecutive frames to demonstrate the facial temporal inconsistency between fake frames (FFD), even when the two frames appear similar at the image level.
More demonstrations can be seen in the supplementary.
} 
\label{fig:artifacts} 
\end{figure}

\section{Introduction}
\label{sec:intro}

Current deepfake technology enables effortless manipulation of facial identities in video content, bringing many impressive applications in fields such as movie-making and entertainment.
However, this technology can also be misused for malicious purposes, including violating personal privacy, spreading misinformation, and eroding trust in digital media.
Hence, there is a pressing need to establish a reliable system for detecting deepfake videos.

Most prior deepfake detectors~\cite{zhou2017two,li2018exposing,rossler2019faceforensics++,yang219exposing} perform well within the same dataset. However, they often fail to generalize well in cross-dataset scenarios where training and testing data distributions differ.
Training with synthetic (blending) images appears to be one of the most effective solutions to this problem, as evidenced by \cite{li2020face,shiohara2022detecting}, which encourages detectors to learn generic representations (\eg, blending boundaries and inconsistencies between inner and outer faces) for deepfake detection.
Given the notable success of image-level blending, we ask: \textit{what about video-level blending? Can this method also be applied to learn general forgery artifacts in the video domain?} (\textbf{Research-Question-1}).

To this end, we perform a comprehensive examination of current realistic forgery videos and discover an inconsistency in the facial organs between forgery frames (as illustrated in Fig.~\ref{fig:artifacts}).
We term this phenomenon as the \textit{Facial Feature Drift (FFD)}. As illustrated in Fig.~\ref{fig:artifacts}, even when two consecutive frames appear relatively static (possibly without other unrelated effects), subtle unnatural drift and flickering can be observed within facial features (\eg, eyes, nose, \etc). 
This observation provides reasonable evidence that unnatural forgery artifacts are indeed present in the temporal domain of manipulated videos.

Intuitively, the root cause of FFD can be attributed to the frame-by-frame face-swapping operation employed by existing deepfake algorithms. 
To understand the underlying reasons, we argue that the primary factor is the randomness during the generation process of fake faces.
Deepfake algorithms typically create a fake frame by using DNNs to generate a counterfeit inner face. Subsequently, they perform a blending operation to integrate this generated face back into the original background.
Given this context, it is important to note that the inner face between different frames within a single generated video may exhibit variations, as the DNN is not a deterministic function, thereby resulting in inconsistencies during the generation process of fake faces.

\begin{figure*} %H为当前位置，!htb为忽略美学标准，htbp为浮动图形
\centering %图片居中
\includegraphics[width=0.8\textwidth]{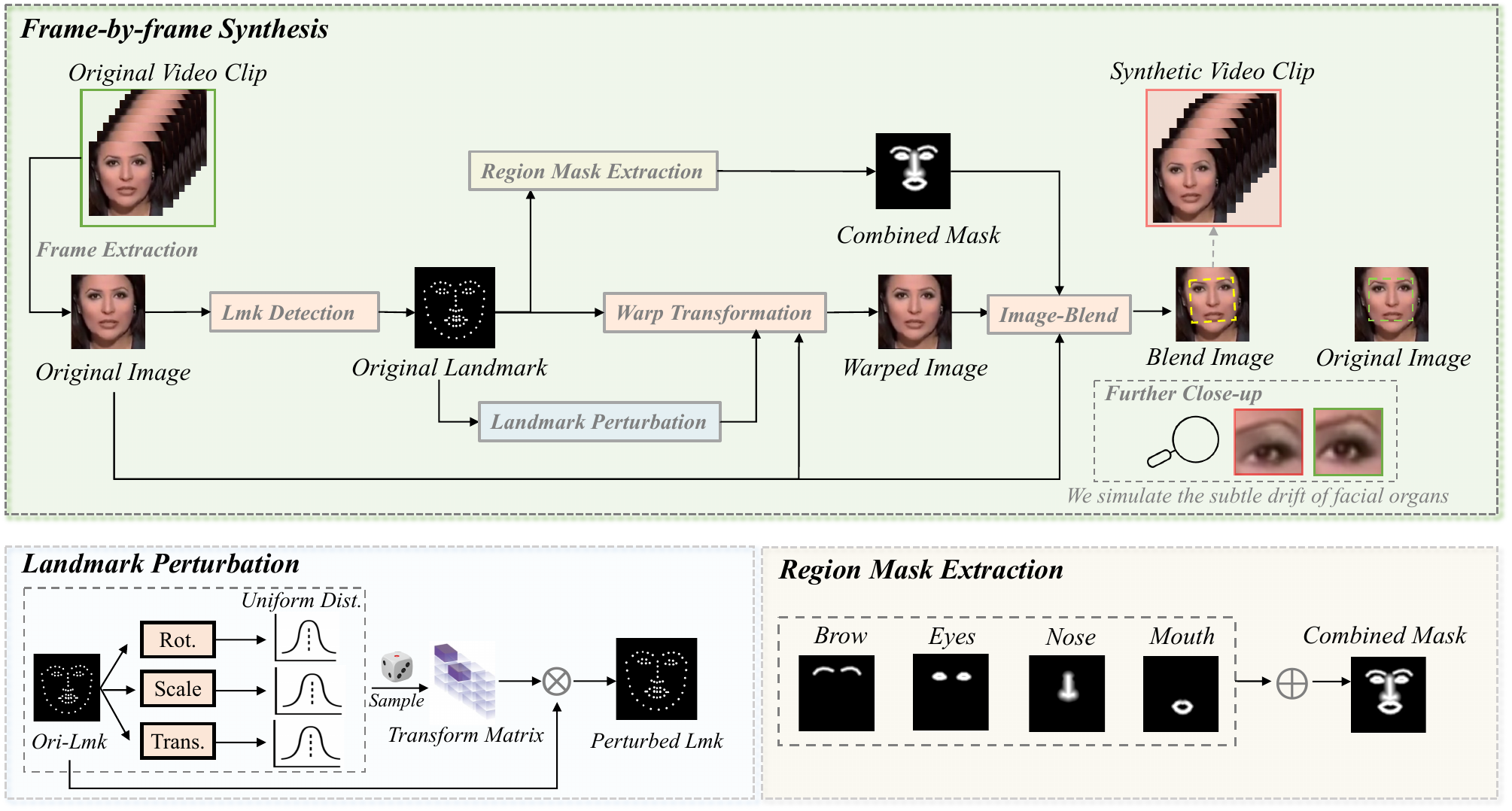} 
\caption{ 
The overall pipeline of the proposed video-level blending method (VB). The whole process involves repeatedly performing \textbf{Frame-by-Frame Synthesis} for a video clip. Two main steps in the frame-by-frame synthesis are \textbf{Landmark Perturbation} and \textbf{Region Mask Extraction}, where the former is designed to add random perturbation to the given facial landmarks and the latter is to extract the mask of each facial organ. The detailed algorithms can be seen in the text.
} 
\label{fig:videoblending} 
\end{figure*}

To effectively encourage the detection model to capture FFD in the given deepfake, we introduce a novel \textbf{V}ideo-level \textbf{B}lending data called \textbf{VB} to simulate and reproduce FFD. The overall creation pipeline of VB is illustrated in Fig.~\ref{fig:videoblending}. 
Specifically, we generate VB by first applying warping operations separately to facial organs and then performing image-level self-blending to generate one synthesis image. The process is repeatedly conducted for all frames within one clip to create a set of synthesis images (video-level pseudo-fakes). VB can serve as the hard negative samples for mining more general discriminative features for detection.
But why do we specifically focus on facial features? This is mainly because regions, such as skin, are low-level areas that can be sensitive to random perturbations (\eg, compression), while facial features typically possess a more robust high-level layout. Also, we can observe more obvious temporal inconsistency within the facial organs in Fig.~\ref{fig:artifacts}.

Additionally, as suggested by \cite{zheng2021exploring,wang2023altfreezing}, naively training a temporal detector may result in the model only focusing on easier types of artifacts and overlooking more challenging ones (\textit{e.g.,} learning spatial only). \textit{So how do we jointly learn temporal and spatial features without heavily relying on one?} (\textbf{Research-Question-2}). Also, \textit{as videos are naturally more complex than images with multiple frames to be computed, how do we address the efficiency issue?} (\textbf{Research-Question-3}).
Instead of designing a new video architecture from scratch, we propose an alternative approach: to ``convert" a pretrained image model into a video model, enabling the model to learn additional temporal information. 
Specifically, we meticulously design a lightweight, plug-and-play Spatiotemporal Adapter (StA) that equips an image model (\eg, ViT or CNN) with spatiotemporal reasoning capabilities.
Throughout the whole training process, only the added adapters are updated, which significantly reduces the training cost and allows us to leverage the powerful generalization capabilities of SoTA image models (\textit{e.g.,} CLIP).

Overall, this work addresses three key challenges in detecting deepfake videos (mentioned in \textbf{Research Questions}). Our main contributions are two-fold as follows.

\begin{itemize}
    
    \item We discover a novel and previously underexplored forgery artifact in the temporal domain. After careful examination of current forgery videos, we observe that there exists a subtle inconsistent artifact in the location and shape of facial organs between frames. We term it \textbf{F}acial \textbf{F}eature \textbf{D}rift (FFD), which is potentially produced within the process of frame-by-frame face-swapping operation. To simulate this, we propose a \textbf{V}ideo-level \textbf{B}lending (VB) to aid the model in learning general temporal forgery features.

    \item We propose a plug-and-play adapter for learning both spatial and temporal forgery artifacts. By solely training the adapters for fine-tuning, we are allowed to leverage the powerful generalization ability of SoTA image models (\eg, CLIP).
\end{itemize}

\section{Related Work}
% Generally, deepfake detection can be classified into two domains: image forgery detection and video forgery detection. We will briefly introduce them separately. Also, we will discuss the related works based on data synthesis (image-level and video-level blending).

\subsection{Deepfake Image Detection}
Conventional image detectors~\cite{nguyen2019capsule, rossler2019faceforensics++} typically focus on developing optimal CNN architectures (\eg, MesoNet, CapsuleNet). However, these methods often overlook the details present in the frequency domain of fake images, such as compression artifacts. To this end, several works~\cite{qian2020thinking,gu2022exploiting} utilize frequency information to improve the performance of detectors. 
Other notable directions are focusing on some specific representations, \textit{i.e.,} forgery region location~\cite{nguyen2019multi}, 3D decomposition~\cite{zhu2021face}, disentanglement~\cite{liang2022exploring,yan2023ucf}, erasing technology~\cite{wang2021representative}, theoretic information~\cite{sun2022information}, and attentional networks~\cite{dang2020detection,wang2022m2tr}. 
These works typically utilize designed algorithms for detecting deepfakes from the view of representation learning.
Another important direction of image forgery detection is data synthesis (image blending). We will discuss it separately in another subsection.

\begin{table*}[t!]
  \centering
   \belowrulesep=0pt
\aboverulesep=0pt
  \resizebox{0.85\textwidth}{!}{%
    \begin{tabular}{c|c|c|c|c}
      \toprule
      \multirow{2}{*}{Methods} & \multicolumn{2}{c|}{Blending Mode} & \multicolumn{2}{c}{Introduced Artifacts} \\
      \cmidrule{2-5}
      %  & \multirow{2}{*}{Image Level}  & \multirow{2}{*}{Video Level} & \multicolumn{2}{c}{Face Inconsistency} \\  
      % \cmidrule{4-5}
      %  &  &  & Face Level & Region Level \\ 
       & Image Level  & Video Level & Face Level & Region Level \\ 
      \midrule
      Face Xray \cite{li2020face}, PCL \cite{zhao2021learning}, SBI \cite{shiohara2022detecting}, OST~\cite{chen2022ost} & \checkmark &  & \checkmark &  \\
      SLADD \cite{chen2022self}  & \checkmark  &   &   & \checkmark  \\
      SeeABLE~\cite{larue2023seeable}                                        &                \checkmark                 &                   &            & \checkmark \\
     
       \midrule 
      \textbf{Ours} &            &\checkmark                                 &                   & \checkmark      \\
      \bottomrule
    \end{tabular}}
  \caption{Comparison of our data synthesis method and other similar methods. Ours clearly differs from previous methods in terms of the level at which blending modes are focused (image-level vs. video-level) and the level at which artifacts are introduced (face-level vs. region-level).
  }
  \label{tab:differences}
\end{table*}

\begin{table*}[h]
\small
\centering
\resizebox{\textwidth}{!}{%
\begin{tabular}{ll}
\hline
Symbol & Definition \\ \hline
    $\mathbf{I}$        & Original image/frame  from the original video.           \\
    $\mathbf{I}'$        & Final blended image/frame.           \\
    $r,R$        & Facial region and set of specific facial regions, \textit{i.e.,} $R = \{r|r\in\{\text{eyes, eyebrows, nose, mouth}\}\}$.           \\
    $\mathbf{I}'_r$        & Blended image/frame about region $r$.           \\
    $\mathbf{I}_{\text{warp}}$        & Warped version of $\mathbf{I}$ obtained through affine transformations.           \\
    $\mathbf{L}_z$        & Set of landmarks defining region $z, z\in R\bigcup\{R\}$.           \\
    $\mathbf{L}^*_R$        & Set of landmarks defining region $R$ after perturbation.           \\
    $u_r$        & Maximum rotation angle for affine transformations.           \\ 
    $u_s$        & Maximum scaling factor for affine transformations.          \\
    $u_t$        & Maximum translation distance for affine transformations.         \\
    $\mathbf{M}_r(x, y)$        & Mask for region $r$, indicating the blending weight at pixel $(x, y)$.           \\
    $\text{dist}(x, y, \mathbf{L}_r)$        & \makecell[l]{Signed distance function calculating the shortest distance from a pixel $(x, y)$ to the closest\\ landmark in $\mathbf{L}_r$.}           \\
    $\text{fdist}_r$        & Predefined fall-off distance for the mask of region $r$, controlling the softness of mask edges.           \\
    $\alpha_r$        & Blending weight for region $r$, determining its influence in the final composite image.           \\
    $\text{clip}(x, min, max)$        & Function that limits $x$ within the range $[min, max]$.           \\\hline
\end{tabular}%
}
\caption{Variables and functions definitions for the proposed VB.}
\label{tab:definition}
\end{table*}

\subsection{Deepfake Video Detection}
There are parts of works~\cite{zhou2017two} that focus on temporal artifacts. 
An early work~\cite{haliassos2021lips} (in 2021) proposes to leverage the temporal inconsistency of the mouth movement.
This work utilizes a pre-trained model to learn the natural lip movement as prior on a lip-reading dataset.
FTCN~\cite{zheng2021exploring} explores directly training a fully temporal 3D ConvNets with an attached temporal Transformer. However, detecting without spatial information may harm the generalization capability.
STIL~\cite{gu2021spatiotemporal} considers both
spatial and temporal inconsistency and designs a spatiotemporal inconsistency Learning framework for deepfake video detection. RealForensics~\cite{haliassos2022leveraging} introduces audio information and leverages self-supervised learning for representation learning. 
Very recently, wang \etal \cite{wang2023altfreezing} propose a new learning method that alternatively trains and freezes the spatial and temporal convolutional layers in a 3D CNN (\eg, I3D R50). The result shows the importance of leveraging both spatial and temporal features for detecting deepfakes.

\subsection{Deepfake Detectors Based on Data Synthesis}
\par \noindent \textbf{Image-level Data Synthesis.}
One effective approach in deepfake detection is to use synthetic data (or blending data/images) for training. 
This strategy encourages models to learn generic representations for detecting deepfakes, such as blending boundaries and inconsistency between the inner and outer faces~\cite{shiohara2022detecting}.
For instance, in the early stages, FWA~\cite{li2018exposing} employs a self-blending strategy by applying image transformations (\textit{e.g.,} down-sampling) to the facial region and then warping it back into the original image. This process is designed to learn the wrapping artifacts during the deepfake generation process. 
Another noteworthy contribution is Face X-ray~\cite{li2020face}, which explicitly encourages detectors to learn the blending boundaries of fake images. 
Furthermore, SLADD~\cite{chen2022self} introduces an adversarial method to dynamically generate the most challenging blending choices for synthesizing data.
Rather than swapping faces between two different identities, a recent art, SBI~\cite{shiohara2022detecting}, proposes to swap with the same person's identity to reach a high-realistic face-swapping.
Given the notable generalization performance for detecting image forgery by using synthetic data, it remains an open question whether and how video-level synthetic data can be effective.

\par \noindent \textbf{Video-level Data Synthesis.}
Compared to image-level data synthesis (blending), very little works pay attention to designing a suitable video-level blending method.
To the best of our knowledge, we identify one notable work~\cite{wang2023altfreezing} that attempts to design video-level data synthesis by using techniques such as dropping/repeating one frame within one clip. 
However, it is possible that these common operations might not 
explicitly represent the general forgery pattern.
Recall that in the image domain, we use blending technology to synthesize the forgery artifact since the blending artifact commonly exists across different forgeries, generally.
So there are two main challenges: (1) What kinds of features generally exist in the temporal domain? (2) How to use blending to simulate it like the image data synthesis?
We will show our video-level method in the following section.

\section{Method}

\subsection{Notation}
We first introduce some important notations and the detailed summary can be found in supplementary.
Suppose we have a frame $\mathbf{I}$ from the original video. We definite that the facial region $R$ contains several specific facial regions $r$ such as eyes, eyebrows, nose, and mouth. Let $\mathbf{L}_z$ be the landmark of region $z$, where $z$ can be a specific facial region $r$ or the whole facial region $R$. To transform landmarks, where transformation parameters are sampled from the uniform distribution, we set 3 hyperparameters $u_r, u_s, u_t$ to be the maximum rotation angle, scaling factor, and translation distance, respectively. After landmark perturbation, we will derive a new landmark $\mathbf{L}^*_R$ of the facial region $R$ and the corresponding warped image $\mathbf{I}_{\text{warp}}$. We consider the mask $\mathbf{M}_r(x, y)$ for region $r$, indicating the blending weight at pixel $(x, y)$. At last, we obtain the final blended frame $\mathbf{I}'$ which is composed of $\mathbf{I}'_r$: blended frame about region $r$, and $\alpha_r$ is used to control the blending weight for region $r$. To better illustrate our method, we denote $\text{dist}(x, y, \mathbf{L}_r)$ as a signed distance function calculating the shortest distance from a pixel $(x, y)$ to the closest landmark in $\mathbf{L}_r$, $\text{fdist}_r$ as a predefined fall-off distance for the mask of region $r$ and $\text{clip}(x, min, max)$ as a function that limits $x$ within the range $[min, max]$.

\subsection{Video-level Data Synthesis}

\begin{algorithm}[H]
\small
\caption{Algorithm of VB}
\label{alg:vb}

\renewcommand{\algorithmicrequire}{\textbf{Input:}}
\renewcommand{\algorithmicensure}{\textbf{Output:}}
\begin{algorithmic}[1] %[1] enables line numbers
\REQUIRE Face image $\mathbf{I}$; Landmarks $\mathbf{L}_R$; Fall\_off distance $\text{fdist}_r, r\in R$; Upper bound of perturbation parameters $u_i,i\in\{r,s,t\}$\\
% \textbf{Parameter}: Encoder E with $\Theta$, Classifier $\mathscr{F1}$ with w1, Classifier $\mathscr{F2}$ with w2, epoch $ T_k$  batch size $B$\\
\ENSURE Blended image $\mathbf{I}'$\\

    \STATE \textit{$\triangleright$} Step 1: \textbf{Warp transformation}
    \STATE Perturbation parameters $\sim U_{[0, u_i]}$
    \STATE Derive affine matrix $\mathbf{A}$
    \STATE $\mathbf{I}_{\text{warp}} \gets \mathbf{A}\mathbf{I}$
    % \STATE shuffle training set $D_f$,$D_g$.
        \FOR{$r$ in $R$}
            % \STATE \COMMENT{Stage 1: Non-face sample selection}
        \STATE \textit{$\triangleright$} Step 2: \textbf{Mask Generation} 
        \STATE $\mathbf{M}_r \gets 1 - \text{clip}\left(\frac{\text{dist}(\mathbf{I}, \mathbf{L}_r)}{\text{fdist}_r}\right)$
        
        \STATE \textit{$\triangleright$} Step 3: \textbf{Dynamic Blending}
        \STATE $\mathbf{I}'_r \gets (\mathbf{M}_r \cdot \mathbf{I}_{\text{warp}}) + ((1 - \mathbf{M}_r) \cdot \mathbf{I})$
        \ENDFOR
    \STATE \textit{$\triangleright$} Step 4:\textbf{Composite Images}
    \STATE $\mathbf{I}' \gets \sum_{r \in R} \alpha_r \cdot \mathbf{I}'_r$
    \RETURN $\mathbf{I}'$
\end{algorithmic}
\end{algorithm}

Although contemporary deepfake generation algorithms are adept at producing realistic forgeries on a frame-by-frame basis, they frequently overlook the consistency between frames. This oversight results in temporal artifacts like flickering and discontinuity. A common temporal artifact known as Facial Feature Drift (FFD) arises from the frame-by-frame process of face-swapping algorithms. This artifact is characterized by inconsistencies in the placement and contours of facial features from one frame to the next. To exploit the generalization potential of FFD for video-level deepfake detection, we introduce a novel synthetic training data, VB: simulating the Facial Feature Drift phenomenon through landmark perturbation and region mask extraction on facial organs across video frames. The self-image blending process involves several steps, including warped image generation, region mask extraction, dynamic blending, and composite image formation, each contributing to the final blended video frame. Note that perturbation parameters $u_r,u_s$, and $u_t$ introduce variability in the appearance of facial regions across frames, contributing to the realism of the video by mimicking natural movements and expressions.

% \begin{itemize}
%     \item $I$: Original image/frame from the video.
%     \item $I'$: Final blended image/frame.
%     \item $R$: Set of specific facial regions $\{r|r\in\{\text{eyes, eyebrows, nose, mouth}\}\}$.
%     % \item $r$: One of specific facial regions $r\in R$.
%     \item $I'_r$: Blended image/frame about region $r$.
%     \item $I_{\text{warp}}$: Warped version of $I$ obtained through affine transformations.
%     \item $L_z$: Set of landmarks defining region $z, z\in R\bigcup\{R\}$.
%     \item $L^*_R$: Set of landmarks defining region $R$ after perturbation.
%     % \item $L_r$: Set of landmarks defining region $r$.
%     \item $u_i$: Upper bound of the parameter to control the perturbation, $i \in \{\text{r},\text{s},\text{t}\}$. (r:rotation; s:scaling; t:translation)
%     \item $M_r(x, y)$: Mask for region $r$, indicating the blending weight at pixel $(x, y)$.
%     \item $\text{dist}(x, y, L_r)$: Signed distance function calculating the shortest distance from a pixel $(x, y)$ to the closest landmark in $L_r$.
%     \item $\text{fdist}_r$: Predefined fall-off distance for the mask of region $r$, controlling the softness of mask edges.
%     \item $\alpha_r$: Blending weight for region $r$, determining its influence in the final composite image.
%     \item $\text{clip}(x, min, max)$: Function that limits $x$ within the range $[min, max]$.
% \end{itemize}

\par \noindent \textbf{Landmark perturbation.}
Our goal is to manifest FFD, that is perturbation of landmarks region. Assuming that the perturbation parameter follows the uniform distribution $U_{[0, u_i]}, i \in \{\text{r},\text{s},\text{t}\}$, we sample parameters for current images perturbation. And we can derive transformed landmarks $\mathbf{L}^*_R=\mathbf{A}\mathbf{L}_R$, where $\mathbf{A}$ is the affine transformation matrix corresponding to sampled parameters. As a result, warp transformation can be written as: 
$\mathbf{I}_{\text{warp}}=\mathbf{A}\mathbf{I}.$

\begin{figure*}[!t] %H为当前位置，!htb为忽略美学标准，htbp为浮动图形
\centering %图片居中
\includegraphics[width=0.85\textwidth]{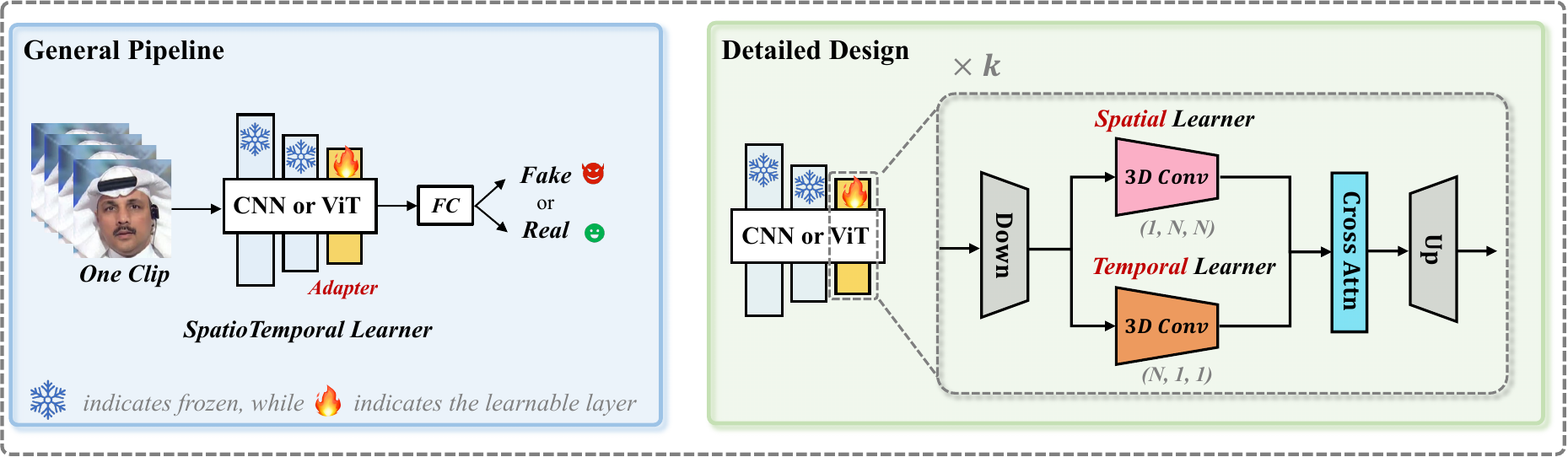} 
\caption{
The overall pipeline of the proposed adapter-based strategy.
We propose a novel and efficient adapter-based method that can be plug-and-play inserted into any SoTA image detector.
} 
\label{fig:adapter} 
\end{figure*}

\par \noindent \textbf{Region Mask Extraction.}
Our method focuses on facial regions, and thus requires a mask to keep other regions the same while changing regions of interest. At the same time, we hope that the closer to the landmark, the more obvious the traces of perturbation will be. For each facial region $r$ with landmarks $\mathbf{L}_r$, the generation of the mask $\mathbf{M}_r$ is formulated as:

\begin{equation}
    \mathbf{M}_r(x, y) = 1 - \text{clip}\left(\frac{\text{dist}(x, y, \mathbf{L}_r)}{\text{fdist}_r}, 0, 1\right), r\in R.
\end{equation}

This equation ensures that the mask $\mathbf{M}_r$ smoothly transitions from 1 (fully included in blending) to 0 (excluded from blending) as the distance from the landmarks increases, up to a maximum defined by $\text{fdist}_r$. The hyper-parameter $\text{fdist}_r$ defines the farthest range that will be included in blending.

% \begin{algorithm}[H]
% \small
% \caption{Algorithm of Training}
% \label{alg:training}

% \renewcommand{\algorithmicrequire}{\textbf{Input:}}
% \renewcommand{\algorithmicensure}{\textbf{Output:}}
% \begin{algorithmic}[1] %[1] enables line numbers
% \small
% \REQUIRE Face image clip $\mathbf{x}$; Clip labels $\mathbf{y}$\\
% \STATE \textbf{PE}: Patchify and embed layers
% \STATE \textbf{TE}: Temporal encoder
% \STATE \textbf{SE}: Spatial encoder
% \STATE \textbf{FC}: Fully connected layers
% % \ENSURE Prediction $\hat{y}$\\

%     % \STATE shuffle training set $D_f$,$D_g$.
%         \FOR{$\mathbf{x}$,$\mathbf{y}$ in Dataloader}
%         \STATE \textit{$\triangleright$} Step 1: \textbf{Embedding}
%         \STATE $\mathbf{x} \gets \textbf{PE}(\mathbf{x})$
%         \STATE \textit{$\triangleright$} Step 2: \textbf{Temporal feature} 
%         \STATE $\mathbf{e}_t \gets \textbf{TE}(\mathbf{x})+\mathbf{x}$
%         \STATE $\mathbf{e}_t \gets \text{Transpose}(\mathbf{e}_t)$
%         \STATE \textit{$\triangleright$} Step 3: \textbf{Spatial feature}
%         \STATE $\mathbf{e}_s \gets \textbf{SE}(\mathbf{e}_t)+\mathbf{e}_t$
%         \STATE \textit{$\triangleright$} Step 4: \textbf{Prediction}
%         \STATE $\hat{\mathbf{y}} \gets \textbf{FC}(\mathbf{e}_s)$
%         % \STATE \textit{$\triangleright$} Step 4: \textbf{Compute loss and backward}
%         \STATE $\text{loss} \gets \text{CrossEntropy}(\hat{\mathbf{y}}, \mathbf{y})$
%         \STATE loss.backward()
%         \ENDFOR

% \end{algorithmic}
% \end{algorithm}

\par \noindent \textbf{Dynamic Blending.}
As we set various masks for different regions, we first create a blended image for each special region. Specifically, the dynamic blending process for a region $r$ is described by the equation:

\begin{equation}
\mathbf{I}'_r = (\mathbf{M}_r \cdot \mathbf{I}_{\text{warp}}) + ((1 - \mathbf{M}_r) \cdot \mathbf{I}), r\in R.  
\end{equation}

Here, $\mathbf{I}_{\text{warp}}$ is derived from $\mathbf{I}$ by applying an affine transformation that introduces perturbations in rotation, scale, and translation, simulating natural movements. $\mathbf{M}_r$ dictates how much of $\mathbf{I}_{\text{warp}}$ versus $\mathbf{I}$ is used at each pixel, achieving a blended effect in the region $r$.

\par \noindent \textbf{Composite Image Formation.}
Finally, VB image $\mathbf{I}'$ can be obtained through compositing blended images of different facial regions. The composite process is formulated as:

\begin{equation}
\mathbf{I}' = \sum_{r \in R} \alpha_r \cdot \mathbf{I}'_r, \quad\text{where} \sum_{r \in R} \alpha_r = 1.
\end{equation}

Each region $r$ contributes to $\mathbf{I}'$ based on its blending weight $\alpha_r$, allowing for the balanced integration of all facial features into the cohesive final image.
After generating a VB frame corresponding to each frame of the video, we synthesize VB frames into a new VB video.

\subsection{Spatiotemporal Adapters (StA)}
A video can be naturally represented as a stack of image frames.
One challenging problem is to capture both spatial and temporal forgery artifacts without heavily relying on one type of artifact.
Previous works generally follow two mainstream frameworks: (1) Utilizing 3D CNN (\eg, I3D R50) for capturing spatiotemporal information from forgery videos~\cite{zheng2021exploring,wang2023altfreezing}; (2) Utilizing two-stream architecture to learn the spatial and temporal features in two separate branches~\cite{masi2020two,zhou2017two}.
These two paradigms typically involve fully fine-tuning for both spatial and temporal modules. 
However, training all layers of both spatial and temporal modules will inevitably introduce a large number of learnable parameters, making the optimization and training procedure challenging.
On the other hand, the image model (especially ViT-based image SoTA) is developing rapidly. It is well-known that these image SoTAs have achieved notable performance in several downstream computer vision tasks (\eg, multi-modality~\cite{ganz2024question}).
Therefore, \textit{can we leverage the power of these SoTA image models to effectively and efficiently finetune the face forgery detection task?}
The key challenge is to perform the temporal forgery feature modeling in a given image model.

To this end, we proposed a lightweight adapter for spatiotemporal feature finetuning.
Our proposed framework is simple but efficient and effective (shown in Fig.~\ref{fig:adapter}).
Differing from previous frameworks for full-parameters fine-tuning, in our framework, only the lightweight adapter is learnable during the training process.
The adapter contains two 3D-Conv layers for spatial and temporal modeling, followed by a cross-attention layer to extract the high-level relationship between them.
The 3D-Conv for spatial modeling is with the kernel size of $(1, N, N)$, capturing the spatial-only information, while the 3D-Conv for temporal is $(N, 1, 1)$, for temporal-only modeling.

Formally, given a video clip $\mathbf{x}_{k}^{in}$ $\in$ $\mathbb{R}^{T\times H\times W\times C}$ (T frames within the same video), obtained from the $l$-th stage of the pretrained image backbone. Where $C$ is the channels of feature maps, $H \times W$ for the spatial size, and $T$ represents the number of frames within the video clip.
The spatial and temporal features can be computed as follows:

\begin{equation}
\begin{aligned}
\mathbf{e}_{k}^{s} &= \operatorname{3DConv_s}\left(\mathbf{W_{down}} \cdot \mathbf{x}_{k}^{in} \right), \\
\mathbf{e}_{k}^{t} &= \operatorname{3DConv_t}\left(\mathbf{W_{down}} \cdot \mathbf{x}_{k}^{in} \right),
\end{aligned}
\end{equation}
where $\mathbf{W_{down}}$ $\in$ $\mathbb{R}^{C \times C'}$. $\operatorname{3DConv_s}$ and $\operatorname{3DConv_t}$ represent the 3D-Conv layers for spatial and temporal modeling. Specifically, we implement three DepthWise 3D-Conv layers with different scales (\textit{i.e.,} 3, 5, 7) to capture multi-scale features.
After obtaining both the spatial ($\mathbf{e}_{k}^{s}$) and temporal feature maps ($\mathbf{e}_{k}^{t}$),
we apply a cross-attention layer to capture the high-level relationship between spatial and temporal features.
First, we flatten the spatial and temporal dimensions and then permute the dimensions to prepare the tensors for multi-head attention: 
$\mathbf{e}_{k}^{s}$ $\rightarrow$ $\mathbf{p}_{k}^{s}$ $\in$ $\mathbb{R}^{(T \times H \times W) \times C}$
and 
$\mathbf{e}_{k}^{t}$ $\rightarrow$ $\mathbf{p}_{k}^{t}$ $\in$ $\mathbb{R}^{(T \times H \times W) \times C}$.
We then apply cross-attention between permuted spatial and temporal features:

\begin{equation}
\begin{aligned}
\mathbf{S2T} = \text{MultiHeadAttention}(\mathbf{p}_{k}^{t}, \mathbf{p}_{k}^{s}, \mathbf{p}_{k}^{s}), \\
\mathbf{T2S} = \text{MultiHeadAttention}(\mathbf{p}_{k}^{s}, \mathbf{p}_{k}^{t}, \mathbf{p}_{k}^{t}),
\end{aligned}
\end{equation}
where $\mathbf{S2T}$ represents the result of integrating spatial information into the temporal dimension, while $\mathbf{T2S}$ represents the result of integrating temporal information into the spatial dimension.

The attention outputs are reshaped back to their original dimensions: $\mathbf{S2T}^{\text{perm}}$ $\rightarrow$ $\mathbf{S2T}^{\text{orig}}$ $\in$ $\mathbb{R}^{B \times C \times T \times H \times W}$ and $\mathbf{T2S}^{\text{perm}}$ $\rightarrow$ $\mathbf{T2S}^{\text{orig}}$ $\in$ $\mathbb{R}^{B \times C \times T \times H \times W}$.
The final output is obtained by averaging the two attention results and adjusting the channels by an Upsampling layer ($up$):

\begin{equation}
\mathbf{x}_{k}^{out} = \frac{1}{2} (\mathbf{S2T}^{\text{orig}} + \mathbf{T2S}^{\text{orig}}) \cdot \mathbf{W_{up}}.
\end{equation}

In the whole training process, only the added modules are updated (\textit{i.e.,} adapters and cross-attention layers, \textit{etc}), significantly reducing the training cost.

\section{Results}

\subsection{Setup}

% \par \noindent \textbf{Dataset.}
% To evaluate the generalization ability of our proposed framework, we conduct experiments on \textbf{eight} commonly used datasets: FaceForensics++ (FF++)~\cite{rossler2019faceforensics++}, Deepfake Detection Challenge (DFDC)~\cite{dfdc}, the preview version of DFDC (DFDCP)~\cite{dfdcp}, DeepfakeDetection (DFD)~\cite{dfd}, Celeb-DF-v2 (CDF-v2)~\cite{li2019celeb}, DeeperForensics (DFo)~\cite{jiang2020deeperforensics}, WildDeepfake (WDF)~\cite{zi2020wilddeepfake}, and FFIW~\cite{zhou2021face}.
% We also evaluate the models on the \textbf{just-released} (in 2024) deepfake dataset DF40~\cite{yan2024df40}, which contains 40 distinct forgery methods. We select \textbf{eight face-swapping methods} generated from the FF++ domain for our evaluation.
% In line with the standard deepfake benchmark~\cite{yan2023deepfakebench}, we utilize the c23 version of FF++ for training and other datasets for testing.
% % It is worth noting that both CDF and DFDC datasets have two versions, originating from similar data sources but differing in data volume. To ensure clarity in our experimental setup, we report results for CDF-v1, CDF-v2, DFDCP (preview version), and DFDC.

\begin{table*}
\centering
\caption{\textbf{Protocol-1:} Cross-dataset evaluations with \textbf{image detectors}. All detectors are trained on FF++\_c23~\cite{rossler2019faceforensics++} and evaluated on other datasets. The best results are highlighted in bold and the second is underlined. * indicates the results are directly cited from their original papers, otherwise the results are cited by \cite{yan2023deepfakebench,cheng2024can,luo2023beyond}.}
\label{tab:table1}
\scalebox{0.8}{
\begin{tabular}{c|c|c|c|c|c|c|c|c|c}
\toprule
\textbf{Detector} & \textbf{Venues} & \textbf{CDF-v2} & \textbf{DFD} & \textbf{DFDC} & \textbf{DFDCP} & \textbf{DFo} & \textbf{WDF} & \textbf{FFIW} & \textbf{Avg.} \\
\midrule
\midrule
F3Net~\cite{qian2020thinking} & ECCV 2020 & 0.789 & 0.844 & 0.718 & 0.749 & 0.730 & 0.728 & 0.649 & 0.743 \\
SPSL~\cite{liu2021spatial} & CVPR 2021 & 0.799 & 0.871 & 0.724 & 0.770 & 0.723 & 0.702 & 0.794 & 0.769\\
SRM~\cite{luo2021generalizing} & CVPR 2021 & 0.840 & 0.885 & 0.695 & 0.728 & 0.722 & 0.702 & 0.794 & 0.767 \\
CORE~\cite{ni2022core} & CVPRW 2022 & 0.809 & 0.882 & 0.721 & 0.720 & 0.765 & 0.724 & 0.710 & 0.762 \\
RECCE~\cite{cao2022end} & CVPR 2022 & 0.823 & 0.891 & 0.696 & 0.734 & 0.784 & 0.756 & 0.711 & 0.779 \\
DCL~\cite{sun2022dual} & AAAI 2022 & 0.882 & 0.921 & \underline{0.750} & 0.769 & 0.975 & 0.776 & 0.863 & 0.851 \\
SLADD*~\cite{chen2022self} & CVPR 2022 & 0.797 & - & 0.772 & - & - & - & -  & - \\
SBI~\cite{shiohara2022detecting} & CVPR 2022 & 0.886 & 0.827 & 0.717 & 0.848 & 0.899 & 0.703 & 0.866 & 0.821 \\
UIA-ViT*~\cite{zhuang2022uia} & ECCV 2022 & 0.889 & - & - & 0.795 & - & - & - & - \\
UCF~\cite{yan2023ucf} & ICCV 2023 & 0.837 & 0.867 & 0.742 & 0.770 & 0.808 & 0.774 & 0.697 & 0.785 \\
SeeABLE*~\cite{yan2023ucf} & ICCV 2023 & 0.873 & - & 0.759 & 0.863 & - & - & - & - \\
% AUNet~\cite{bai2023aunet} & CVPR 2023 & 0.928 & 0.992 & 0.738 & 0.862 & - & - & 0.815 & - \\
IID*~\cite{huang2023implicit} & CVPR 2023 & 0.838 & 0.939 & - & \underline{0.812} & - & - & - & - \\
CFM~\cite{luo2023beyond} & TIFS 2023 & 0.897 & 0.952 & 0.706 & 0.802 & \underline{0.976} & \underline{0.823} & 0.831 & \underline{0.856} \\
LSDA~\cite{yan2023transcending} & CVPR 2024 & \underline{0.898} & \underline{0.956} & 0.735 & \underline{0.812} & 0.892 & 0.756 & \underline{0.879} & 0.847 \\
\midrule
\multirow{2}*{\textbf{Ours}} &\multirow{2}*{-} & \textbf{0.947} & \textbf{0.965} & \textbf{0.843} & \textbf{0.909} & \textbf{0.991} & \textbf{0.848} & \textbf{0.921} & \textbf{0.911} \\
& & \textcolor{cvprgreen}{(↑4.9\%)} & \textcolor{cvprgreen}{(↑0.9\%)} & \textcolor{cvprgreen}{(↑9.3\%)} & \textcolor{cvprgreen}{(↑6.1\%)} & \textcolor{cvprgreen}{(↑1.5\%)} & \textcolor{cvprgreen}{(↑2.5\%)} & \textcolor{cvprgreen}{(↑4.2\%)} & \textcolor{cvprgreen}{(↑5.5\%)} \\
\bottomrule
\end{tabular}
}
\end{table*}

\begin{table*}

  \centering
  \caption{
    \textbf{Protocol-2:} Cross-manipulation evaluations within FF++ domain. All models are trained on FF++ (c23) and tested on other forgeries. $\dagger$ indicates the results are obtained by using the official pre-trained model for evaluation, otherwise the results are cited by \cite{yan2023deepfakebench,cheng2024can}.}
  \label{protocol2}
  \scalebox{0.8}{
  \begin{tabular}{c|c|c|c|c|c|c|c|c|c} 
  \toprule
    \textbf{Detector} & \textbf{UniFace} & \textbf{BlendFace} & \textbf{MobileSwap} & \textbf{e4s} & \textbf{FaceDancer} & \textbf{FSGAN} & \textbf{Inswap} & \textbf{SimSwap} & \textbf{Avg.} \\
    
    \midrule
    \midrule

    RECCE~\cite{cao2022end} & 0.898 & 0.832 & 0.925 & 0.683 & 0.848 & 0.949 & 0.848 & 0.768 & 0.844 \\
    SBI~\cite{shiohara2022detecting} & 0.724 & 0.891 & \textbf{0.952} & \underline{0.750} & 0.594 & 0.803 & 0.712 & 0.701 & 0.766 \\
    UCF~\cite{yan2023ucf} & 0.831 & 0.827 & \underline{0.950} & 0.731 & \underline{0.862} & 0.937 & 0.809 & 0.647 & 0.824 \\
    AltFreezing$\dagger$~\cite{wang2023altfreezing} & \underline{0.947} & \textbf{0.951} & 0.851 & 0.605 & 0.836 & \underline{0.958} & \underline{0.927} & \underline{0.797} & \underline{0.859} \\
    \midrule
    \textbf{Ours} & \textbf{0.960} & \underline{0.906} & 0.946 & \textbf{0.980} & \textbf{0.912} & \textbf{0.964} & \textbf{0.937} & \textbf{0.931} & \textbf{0.942} \\ 
    \bottomrule
  \end{tabular}
  }
\end{table*}

\begin{table}[t!]
   
    \renewcommand\arraystretch{1.25}
    \centering

    \captionof{table}{
        \textbf{Protocol-3:} Cross-dataset evaluations with \textbf{video detectors}. The results of these detectors are directly cited from their original papers.
    }
    \label{tab:cmp_sota}
    \resizebox{0.9\columnwidth}{!}{
        \begin{tabular}{c|c|c|c|c} \toprule
        \textbf{Detector} & \textbf{Venues} & \textbf{CDF-v2} & \textbf{DFDC} & \textbf{Avg.}\\ 
        \midrule
        LipForensics~\cite{haliassos2021lips} & CVPR 2021 & 0.824 & 0.735 & 0.780\\
        FTCN~\cite{zheng2021exploring} & ICCV 2021 & 0.869 & 0.740 & 0.805 \\
        HCIL~\cite{gu2022hierarchical} & ECCV 2022 & 0.790 & 0.692 & 0.741\\
        RealForensics~\cite{haliassos2022leveraging} & CVPR 2022 & 0.857 & 0.759 & 0.808 \\
        LTTD~\cite{guan2022delving} & NeurIPS 2022 & 0.893 & \underline{0.804} & \underline{0.849} \\
        AltFreezing~\cite{wang2023altfreezing} & CVPR 2023 & 0.895 & - & - \\
        TALL-Swin~\cite{xu2023tall} & ICCV 2023 & \underline{0.908} & 0.768 & 0.838 \\
        StyleDFD~\cite{choi2024exploiting} & CVPR 2024 & 0.890 & - & - \\
        NACO~\cite{zhang2024learning} & ECCV 2024 & 0.895 & 0.767 & 0.831 \\
    
        \midrule
        
        \multirow{2}*{\textbf{Ours}} & \multirow{2}*{-} & \textbf{0.947} & \textbf{0.843} & \textbf{0.895}\\
    
        & & \textcolor{cvprgreen}{(↑3.9\%)} & \textcolor{cvprgreen}{(↑3.9\%)} & \textcolor{cvprgreen}{(↑4.6\%)}\\
        
        \bottomrule
      \end{tabular}
    }
  \vspace{-4mm}
    
    \end{table}

\begin{table}

  \centering
  \caption{
  Ablation studies regarding the effectiveness of our proposed plug-and-play strategy (VB + StA). 
  % Some of the results are cited from \cite{xu2023tall}.
  }
  \label{tab:params}
  \scalebox{0.6}{
  \begin{tabular}{c|c|c|c|c|c} 
  \toprule
    \textbf{Detector} & \textbf{\#Params} & \textbf{Temporal} & \textbf{CDF-v2} & \textbf{DFDC} & \textbf{Avg.} \\ 
    \cmidrule(lr){1-1}
    \cmidrule(lr){2-2}
    \cmidrule(lr){3-3}
    \cmidrule(lr){4-4}
    \cmidrule(lr){5-5}
    \cmidrule(lr){6-6}
    I3D-RGB~\cite{haliassos2021lips} & 25M & \checkmark & 0.625 & 0.655 & 0.640 \\
    3D R50~\cite{haliassos2021lips} & 46M & \checkmark & 0.824 & 0.735 & 0.780 \\
    VTN~\cite{gu2021spatiotemporal} & 46M & \checkmark & 0.734 & 0.691 & 0.713 \\
    VidTR~\cite{zheng2021exploring} & 93M & \checkmark & 0.869 & 0.740 & 0.805 \\
    ViViT~\cite{wang2023altfreezing} & 310M & \checkmark & 0.895 & - & - \\ 
    ISTVT~\cite{tong2022videomae} & - & \checkmark & 0.606 & 0.551 & 0.579 \\
    \midrule

    ResNet-34~\cite{he2016deep} & 21M & $\times$ & 0.742 & 0.626 & 0.684 \\
    \textbf{+} \textbf{Ours} & 21M & \checkmark & 0.829 & 0.666 & 0.748 \\
    ViT-b16-21k~\cite{dosovitskiy2020image} & 84M & $\times$ & 0.834 & 0.739 & 0.778 \\
    \textbf{+} \textbf{Ours} & 22M & \checkmark & 0.855 & 0.772 & 0.814 \\
    CLIP-b16~\cite{radford2021learning} & 84M & $\times$ & 0.848 & 0.741 & 0.795 \\
    \textbf{+} \textbf{Ours} & 22M & \checkmark & 0.866 & 0.778 & 0.822 \\
    CLIP-l14~\cite{radford2021learning} & 307M & $\times$ & 0.900 & 0.798 & 0.854 \\
    \textbf{+} \textbf{Ours} & 46M & \checkmark & \textbf{0.947} & \textbf{0.843} & \textbf{0.895} \\ 
    \bottomrule
  \end{tabular}
  }
  \vspace{-4mm}
\end{table}

\par \noindent \textbf{Implementation Details.}
We use the CLIP ViT-l14~\cite{radford2021learning} as the default backbone. We also explore other architectures for experiments, as shown in Tab.~\ref{tab:params}.
All preprocessing and training codebases are aligned to the deepfake benchmark~\cite{yan2023deepfakebench}.
Specifically, we uniformly sample 32 frames from each video during training and randomly select 8 consecutive frames as a clip for each iteration. 
We also employ several widely used data augmentations, such as CutOut, and ImageCompression.

\par \noindent \textbf{Evaluation Metrics.} 
By default, we report the video-level Area Under the Curve (AUC) to compare our proposed method with prior works.
Other evaluation metrics such as Accuracy (Acc.) and Equal Error Rate (EER) are also reported for a more comprehensive evaluation.

\par \noindent \textbf{Evaluation Protocols and Dataset.} 
We adopt three standard protocols to evaluate our models: \textbf{Protocol-1:} cross-dataset evaluation with image methods, \textbf{Protocol-2:} cross-manipulation evaluation on the FF++ data domain, and \textbf{Protocol-3:} cross-dataset evaluation with video methods.
For \textbf{Protocol-1} and \textbf{Protocol-3}, we evaluate the models by training them on FaceForensics++ (FF++)~\cite{rossler2019faceforensics++} and then testing them on seven other deepfake detection datasets. 
Note that FF++ has three different compression versions and we adopt the c23 version for training all methods in our experiments, following most existing works~\cite{yan2023transcending}.
The evaluation datasets include Celeb-DF-v2 (CDF-v2)~\cite{li2019celeb}, DeepfakeDetection (DFD)~\cite{dfd}, Deepfake Detection Challenge (DFDC)~\cite{dfdc}, the preview version of DFDC (DFDCP)~\cite{dfdcp}, DeeperForensics (DFo)~\cite{jiang2020deeperforensics}, WildDeepfake (WDF)~\cite{zi2020wilddeepfake}, and FFIW~\cite{zhou2021face}.
For \textbf{Protocol-2}, we evaluate the models using the latest deepfake dataset DF40~\cite{yan2024df40}, which contains diverse forgery data generated within the FF++ domain. By doing this, we can ensure that although the evaluation involves different forgeries, the data domain remains the same.

\subsection{Generalization Performance Evaluation}
Forgery methods in the real world can be highly diverse. Evaluating our model's performance in diverse scenarios is needed.
We conduct both cross-dataset evaluations (Protocol-1 and Protocol-3) and cross-manipulation tasks, allowing us to determine whether the model can handle both previously unseen data domains and unseen fake methods.
Results in Fig.~\ref{tab:table1}, Fig.~\ref{protocol2}, and Fig.~\ref{tab:cmp_sota} show that our method generally outperforms other models on average, particularly on e4s, Inswap, and SimSwap methods (see Tab.~\ref{protocol2}). 
This demonstrates that our method learns more general features for detection, even for the latest techniques.

\begin{figure*}[!t] %H为当前位置，!htb为忽略美学标准，htbp为浮动图形
\centering %图片居中
\includegraphics[width=1.0\textwidth]{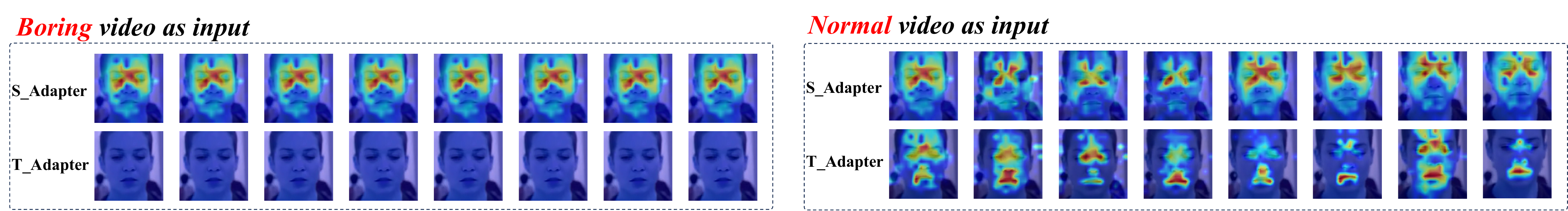} 
\caption{GradCAM for demonstrating the attention regions of S- and T-Adapters.
We create a ``boring" video by repeating a single image into a video sequence.
We show that the T-adapter can capture reasonable temporal-related motion like mouth movement, while the S-adapter's outputs remain constant due to the same input of the boring video.} 
\label{fig:gradcam} 
\end{figure*}

\begin{figure*}[!t] %H为当前位置，!htb为忽略美学标准，htbp为浮动图形
\centering %图片居中
\includegraphics[width=0.8\textwidth]{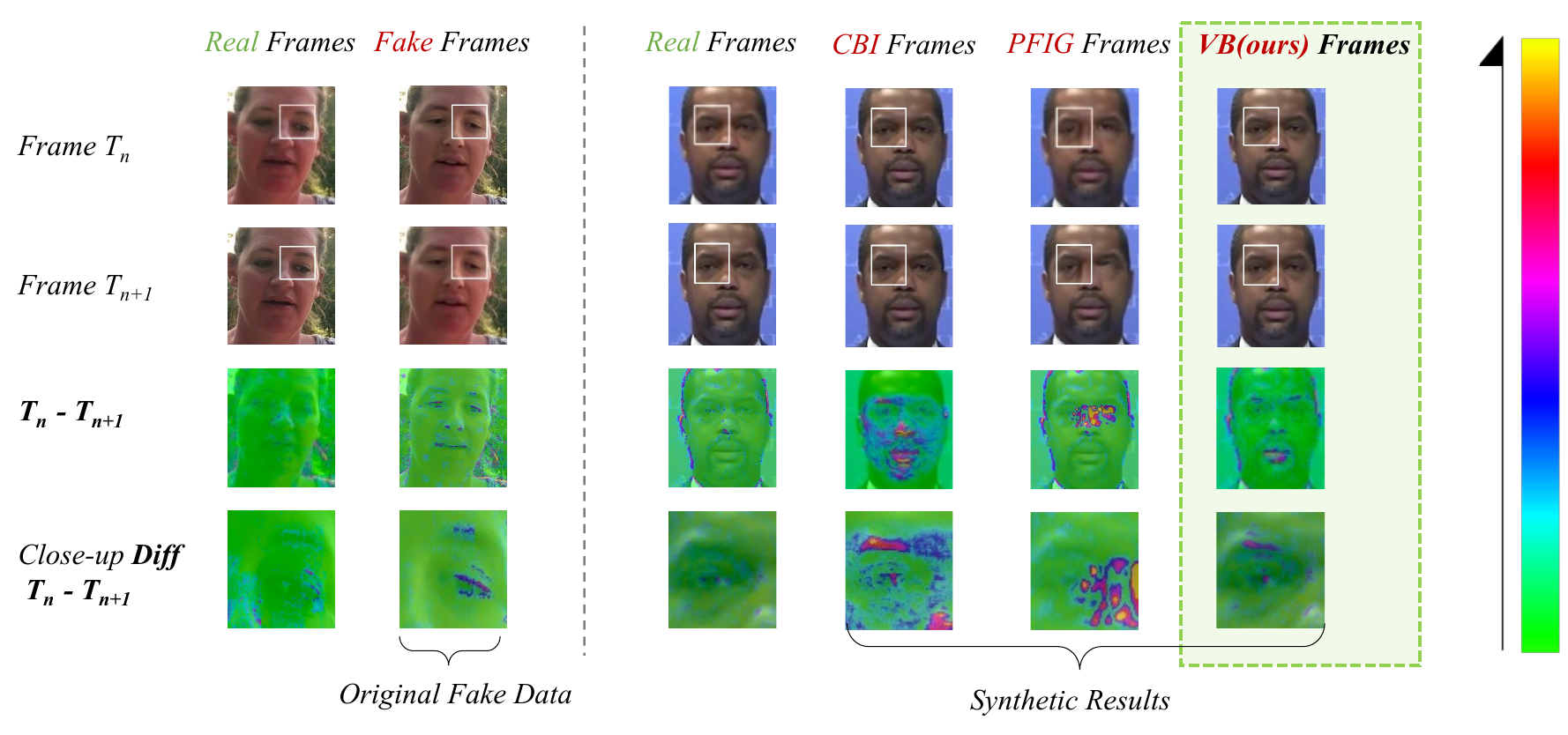} 
\caption{
Comparison of different video-level blending strategies. We consider other two possible solutions to simulate the FFD: (1) CBI, which represents the face-hull blending of two frames with the same video clip; (2) PFIG~\cite{sun2023towards}, which represents the facial-region blending of the different videos.
Our simulation method VB shows the most similar result to the original FFD in the fake data.
Best viewed in color.
} 
\label{fig:vb_comp} 
\end{figure*}

\begin{table}[t!]
    \renewcommand\arraystretch{1.25}
    \centering
    \captionof{table}{
    Ablation studies regarding the effectiveness of StA and VB, separately.  
    We adopt the AUC, Acc, and EER metrics for reporting.
    % All models are trained on the FF++\_c23 dataset and evaluated across various other two mainstream evaluation datasets with metrics presented in the order of AUC~\textbar~Acc~\textbar~EER (the video-level). The average performance (Avg.) across all datasets is also reported.
    }
    \label{table:ablation_all}  
    \resizebox{0.9\columnwidth}{!}{
    \begin{tabular}{c|c|c|c} \toprule
    \multirow{2}*{StA} & \multirow{2}*{VB} & CDF-v2 & WDF\\ 
    \cmidrule(lr){4-4}
    \cmidrule(lr){3-3}
    & & AUC  ~\textbar~  Acc.  ~\textbar~ EER
    &  AUC  ~\textbar~  Acc. ~\textbar~ EER \\
    \midrule
    $\times$ & $\times$ & 0.900 ~\textbar~ 0.739 ~\textbar~ 0.197  & 0.794 ~\textbar~ 0.572 ~\textbar~ 0.249 \\

    \checkmark & $\times$ & 0.926 ~\textbar~ 0.812 ~\textbar~ 0.195  & 0.842 ~\textbar~ 0.654 ~\textbar~ 0.222 \\
    
    \checkmark &  \checkmark & \textbf{0.947} ~\textbar~ \textbf{0.838} ~\textbar~ \textbf{0.176} & \textbf{0.848} ~\textbar~ \textbf{0.702} ~\textbar~ \textbf{0.212} \\    
    
    \bottomrule
  \end{tabular}

    } 

\end{table}

\subsection{Ablation Study}

\paragraph{Effectiveness of our plug-and-play strategy.}
To show the \textit{extensibility} of our method, we apply it to various architectures, including ResNet-34 (CNN) and CLIP (ViT). Results in Tab.~\ref{tab:params} indicate that our proposed method can imbue the original model with \textit{spatiotemporal} modeling abilities. Our method also achieves better results with fewer training parameters, emphasizing the advantages of our design.

\paragraph{Effectiveness of VB and StA.}
To evaluate the impact of the two proposed strategies (VB and StA) separately, we conduct ablation studies on CDF-v2 and WDF.
The evaluated variants include the baseline CLIP-l14 without any further modifications, the baseline with the proposed StA, and our overall framework (StA + VB).
The incremental enhancement in the overall generalization performance with the addition of each strategy, as evidenced by the results in Tab.~\ref{table:ablation_all}, underscores the effectiveness of these strategies. 
% Notably, after applying VB, the Acc improves largely on both CDF-v2 (+13.3\%) and DFDCP (+13.1\%).
% Also, the results indicate the contributions of all proposed strategies to the outcomes. Omitting either of them can lead to a substantial decrease in the overall performance.

\begin{table}

  \centering
  \caption{
    Comparison of Fixed SBI and the proposed VB.
  }
  \label{tab:sbi}
  \scalebox{0.7}{
  \begin{tabular}{c|c|c|c|c|c} 
  \toprule
    \textbf{Detector} & \textbf{Spatial} & \textbf{Temporal} & \textbf{CDF-v2} & \textbf{DFDC} & \textbf{Avg.} \\ 
    \midrule
    Fixed SBI~\cite{haliassos2021lips} & \checkmark & $\times$ & 0.902 & 0.815 & 0.859 \\
    VB (proposed method) & $\times$ & \checkmark & 0.947 & 0.843 & 0.895 \\
    Fixed SBI + VB & \checkmark & \checkmark & 0.951 & 0.846 & 0.899 \\
    \bottomrule
  \end{tabular}
  }
\end{table}

\begin{table}

  \centering
  \caption{
    Ablation studies of the proposed cross-attention.
  }
  \label{tab:cross-attn}
  \scalebox{0.7}{
  \begin{tabular}{c|c|c|c|c|c} 
  \toprule
    \textbf{Fusion Method} & \textbf{CDF-v2} & \textbf{DFDC} & \textbf{WDF} & \textbf{DFo} & \textbf{Avg.} \\ 
    \midrule
    Linear combination ($\mathbf{e}_{k}^{s} + \mathbf{e}_{k}^{t})$ & 0.886 & 0.787 & 0.805 & 0.974 & 0.863 \\
    Cross-Attn ($\mathbf{S2T}^{\text{orig}} + \mathbf{T2S}^{\text{orig}}$) & 0.947 & 0.843 & 0.848 & 0.991 & 0.905 \\
    \bottomrule
  \end{tabular}
  }
  \vspace{-4mm}
  
\end{table}

\vspace{-4mm}

\paragraph{Image blending \textit{vs.} video blending.}
Can image blending method, \textit{e.g.,} SBI, act as a ``video augmentor"? We designed an ablation study to investigate the effectiveness of \textbf{fixed SBI} (without temporal inconsistency) and \textbf{VB} (without spatial artifacts) in video detection.
Results in Table~\ref{tab:sbi} show that SBI achieves lower results than VB, but the combination of both SBI and VB is more effective than VB alone. This suggests that SBI and VB could be complementary, with one focusing on spatial aspects and the other on temporal aspects. This finding may inspire future research on how to effectively utilize both techniques to improve generalization in video detection.

\vspace{-4mm}

\paragraph{Effectiveness of the proposed cross-attention module.}
By our experiments, we surprisingly found that the cross-attention module is significantly important in our StA architecture (see Tab.~\ref{tab:cross-attn}). This might indicate that the spatial and temporal information can be benefited to each other, potentially in a high-level manner (not linear).

\vspace{-4mm}

\paragraph{Evidence for the proposed spatiotemporal adapters can indeed learn spatial and temporal clues?}
To verify it with visualizations, we create a (boring) video by repeating a single image into a video sequence.
We then pass the input to our spatial (S) and temporal (T) adapters (two pathways with different kernel sizes of 3D Convs). 
As expected, Fig.~\ref{fig:gradcam} shows that the outputs of S remain constant (same inputs), while the outputs of T are zero (no change in time).
Plus, for a normal video as input, T\_adapter can capture reasonable temporal-related motion, \textit{e.g.,} mouth movement.

\paragraph{Comparison of different video-level blending solutions.}
To reproduce and simulate the FFD artifact, we propose a novel video-level blending data called VB.
We also attempt to use other choices for implementation. In this subsection, we discuss the effectiveness of different video-level blending solutions for reproducing the FFD. Here, we discuss three potential solutions: (1) CBI (blend with other frames within one clip); (2) PFIG, the same method proposed in \cite{sun2023towards} that performs the facial-region blending of two frames from different videos.
\textbf{First}, we consider CBI, which performs the blending operation with another frame within the same clip. This method could introduce the inconsistent time shift of the facial region between frames. However, as we can see in Fig.~\ref{fig:vb_comp}, the inconsistency between frames is also too obvious, which does not align with the original fake (FFD).
\textbf{Second}, we also consider PFIG~\cite{sun2023towards}, which performs blending between two frames (one real and one fake). PFIG randomly chooses one facial region of the fake and blends it into the real. Here, to simulate the FFD, we selectively choose a facial region from another frame within the same clip and blend it back into the real to generate the synthetic data. As we can see in Fig.~\ref{fig:vb_comp}, the blending region is mainly within a specific region (\textit{e.g.,} eyes), not aligned with the original fake.

\section{Conclusion}
In this paper, we uncover a novel and previously underexplored temporal forgery artifact present across various manipulated videos, which we term \underline{F}acial \underline{F}eature \underline{D}rift (FFD). To simulate this artifact, we propose a novel \underline{V}ideo-level \underline{B}lending data (VB), designed to aid the model in learning more general forgery artifacts within the temporal domain.
We also propose an adapter-based strategy for efficiently and effectively learning both the spatial and temporal forgery artifacts, without being overfitting to one specific type (\textit{e.g.,} learning spatial only).
Extensive experiments on several widely used and latest deepfake datasets verify the effectiveness of the proposed methods.
% We hope that our work will inspire further research into (1) the development of video augmentations (also the approach to combine both image and video blending), and (2) efficient spatiotemporal adapter tunning in the deepfake video detection community.

\clearpage

{\small
\bibliographystyle{ieeenat_fullname}
\bibliography{refer} 

\begin{thebibliography}{57}
\providecommand{\natexlab}[1]{#1}
\providecommand{\url}[1]{\texttt{#1}}
\expandafter\ifx\csname urlstyle\endcsname\relax
  \providecommand{\doi}[1]{doi: #1}\else
  \providecommand{\doi}{doi: \begingroup \urlstyle{rm}\Url}\fi

\bibitem[Cao et~al.(2022)Cao, Ma, Yao, Chen, Ding, and Yang]{cao2022end}
Junyi Cao, Chao Ma, Taiping Yao, Shen Chen, Shouhong Ding, and Xiaokang Yang.
\newblock End-to-end reconstruction-classification learning for face forgery detection.
\newblock In \emph{Proceedings of the IEEE/CVF Conference on Computer Vision and Pattern Recognition}, pages 4113--4122, 2022.

\bibitem[Chen et~al.(2022{\natexlab{a}})Chen, Zhang, Song, Liu, and Wang]{chen2022self}
Liang Chen, Yong Zhang, Yibing Song, Lingqiao Liu, and Jue Wang.
\newblock Self-supervised learning of adversarial example: Towards good generalizations for deepfake detection.
\newblock In \emph{Proceedings of the IEEE/CVF Conference on Computer Vision and Pattern Recognition}, pages 18710--18719, 2022{\natexlab{a}}.

\bibitem[Chen et~al.(2022{\natexlab{b}})Chen, Zhang, Song, Wang, and Liu]{chen2022ost}
Liang Chen, Yong Zhang, Yibing Song, Jue Wang, and Lingqiao Liu.
\newblock Ost: Improving generalization of deepfake detection via one-shot test-time training.
\newblock In \emph{Proceedings of the Neural Information Processing Systems}, 2022{\natexlab{b}}.

\bibitem[Cheng et~al.(2024)Cheng, Yan, Zhang, Luo, Wang, and Li]{cheng2024can}
Jikang Cheng, Zhiyuan Yan, Ying Zhang, Yuhao Luo, Zhongyuan Wang, and Chen Li.
\newblock Can we leave deepfake data behind in training deepfake detector?
\newblock \emph{arXiv preprint arXiv:2408.17052}, 2024.

\bibitem[Choi et~al.(2024)Choi, Kim, Jeong, Baek, and Choi]{choi2024exploiting}
Jongwook Choi, Taehoon Kim, Yonghyun Jeong, Seungryul Baek, and Jongwon Choi.
\newblock Exploiting style latent flows for generalizing deepfake video detection.
\newblock In \emph{Proceedings of the IEEE/CVF Conference on Computer Vision and Pattern Recognition}, pages 1133--1143, 2024.

\bibitem[Dang et~al.(2020)Dang, Liu, Stehouwer, Liu, and Jain]{dang2020detection}
Hao Dang, Feng Liu, Joel Stehouwer, Xiaoming Liu, and Anil~K Jain.
\newblock On the detection of digital face manipulation.
\newblock In \emph{Proceedings of the IEEE/CVF Conference on Computer Vision and Pattern Recognition}, 2020.

\bibitem[Deepfakedetection.(2021)]{dfd}
Deepfakedetection., 2021.
\newblock \url{https://ai.googleblog.com/2019/09/contributing-data-to-deepfakedetection.html} Accessed 2021-11-13.

\bibitem[Dolhansky et~al.(2019)Dolhansky, Howes, Pflaum, Baram, and Ferrer]{dfdcp}
Brian Dolhansky, Russ Howes, Ben Pflaum, Nicole Baram, and Cristian~Canton Ferrer.
\newblock The deepfake detection challenge (dfdc) preview dataset.
\newblock \emph{arXiv preprint arXiv:1910.08854}, 2019.

\bibitem[Dolhansky et~al.(2020)Dolhansky, Bitton, Pflaum, Lu, Howes, Wang, and Ferrer]{dfdc}
Brian Dolhansky, Joanna Bitton, Ben Pflaum, Jikuo Lu, Russ Howes, Menglin Wang, and Cristian~Canton Ferrer.
\newblock The deepfake detection challenge (dfdc) dataset.
\newblock \emph{arXiv preprint arXiv:2006.07397}, 2020.

\bibitem[Dosovitskiy et~al.(2020)Dosovitskiy, Beyer, Kolesnikov, Weissenborn, Zhai, Unterthiner, Dehghani, Minderer, Heigold, Gelly, et~al.]{dosovitskiy2020image}
Alexey Dosovitskiy, Lucas Beyer, Alexander Kolesnikov, Dirk Weissenborn, Xiaohua Zhai, Thomas Unterthiner, Mostafa Dehghani, Matthias Minderer, Georg Heigold, Sylvain Gelly, et~al.
\newblock An image is worth 16x16 words: Transformers for image recognition at scale.
\newblock \emph{arXiv preprint arXiv:2010.11929}, 2020.

\bibitem[Ganz et~al.(2024)Ganz, Kittenplon, Aberdam, Avraham, Nuriel, Mazor, and Litman]{ganz2024question}
Roy Ganz, Yair Kittenplon, Aviad Aberdam, Elad~Ben Avraham, Oren Nuriel, Shai Mazor, and Ron Litman.
\newblock Question aware vision transformer for multimodal reasoning, 2024.

\bibitem[Gu et~al.(2022{\natexlab{a}})Gu, Chen, Yao, Chen, Ding, and Yi]{gu2022exploiting}
Qiqi Gu, Shen Chen, Taiping Yao, Yang Chen, Shouhong Ding, and Ran Yi.
\newblock Exploiting fine-grained face forgery clues via progressive enhancement learning.
\newblock In \emph{Proceedings of the AAAI Conference on Artificial Intelligence}, pages 735--743, 2022{\natexlab{a}}.

\bibitem[Gu et~al.(2021)Gu, Chen, Yao, Ding, Li, Huang, and Ma]{gu2021spatiotemporal}
Zhihao Gu, Yang Chen, Taiping Yao, Shouhong Ding, Jilin Li, Feiyue Huang, and Lizhuang Ma.
\newblock Spatiotemporal inconsistency learning for deepfake video detection.
\newblock In \emph{Proceedings of the 29th ACM international conference on multimedia}, pages 3473--3481, 2021.

\bibitem[Gu et~al.(2022{\natexlab{b}})Gu, Yao, Chen, Ding, and Ma]{gu2022hierarchical}
Zhihao Gu, Taiping Yao, Yang Chen, Shouhong Ding, and Lizhuang Ma.
\newblock Hierarchical contrastive inconsistency learning for deepfake video detection.
\newblock In \emph{Proceedings of the European Conference on Computer Vision}, pages 596--613. Springer, 2022{\natexlab{b}}.

\bibitem[Guan et~al.(2022)Guan, Zhou, Hong, Ding, Wang, Quan, and Zhao]{guan2022delving}
Jiazhi Guan, Hang Zhou, Zhibin Hong, Errui Ding, Jingdong Wang, Chengbin Quan, and Youjian Zhao.
\newblock Delving into sequential patches for deepfake detection.
\newblock \emph{Advances in Neural Information Processing Systems}, 35:\penalty0 4517--4530, 2022.

\bibitem[Haliassos et~al.(2021)Haliassos, Vougioukas, Petridis, and Pantic]{haliassos2021lips}
Alexandros Haliassos, Konstantinos Vougioukas, Stavros Petridis, and Maja Pantic.
\newblock Lips don't lie: A generalisable and robust approach to face forgery detection.
\newblock In \emph{Proceedings of the IEEE/CVF Conference on Computer Vision and Pattern Recognition}, 2021.

\bibitem[Haliassos et~al.(2022)Haliassos, Mira, Petridis, and Pantic]{haliassos2022leveraging}
Alexandros Haliassos, Rodrigo Mira, Stavros Petridis, and Maja Pantic.
\newblock Leveraging real talking faces via self-supervision for robust forgery detection.
\newblock In \emph{Proceedings of the IEEE/CVF Conference on Computer Vision and Pattern Recognition}, pages 14950--14962, 2022.

\bibitem[He et~al.(2016)He, Zhang, Ren, and Sun]{he2016deep}
Kaiming He, Xiangyu Zhang, Shaoqing Ren, and Jian Sun.
\newblock Deep residual learning for image recognition.
\newblock In \emph{Proceedings of the IEEE/CVF Conference on Computer Vision and Pattern Recognition}, pages 770--778, 2016.

\bibitem[Huang et~al.(2023)Huang, Wang, Yang, Ai, Zou, Wang, and Ye]{huang2023implicit}
Baojin Huang, Zhongyuan Wang, Jifan Yang, Jiaxin Ai, Qin Zou, Qian Wang, and Dengpan Ye.
\newblock Implicit identity driven deepfake face swapping detection.
\newblock In \emph{Proceedings of the IEEE/CVF Conference on Computer Vision and Pattern Recognition}, pages 4490--4499, 2023.

\bibitem[Jiang et~al.(2020)Jiang, Li, Wu, Qian, and Loy]{jiang2020deeperforensics}
Liming Jiang, Ren Li, Wayne Wu, Chen Qian, and Chen~Change Loy.
\newblock Deeperforensics-1.0: A large-scale dataset for real-world face forgery detection.
\newblock In \emph{Proceedings of the IEEE/CVF Conference on Computer Vision and Pattern Recognition}, 2020.

\bibitem[Larue et~al.(2023)Larue, Vu, Struc, Peer, and Christophides]{larue2023seeable}
Nicolas Larue, Ngoc-Son Vu, Vitomir Struc, Peter Peer, and Vassilis Christophides.
\newblock Seeable: Soft discrepancies and bounded contrastive learning for exposing deepfakes.
\newblock In \emph{Proceedings of the IEEE/CVF International Conference on Computer Vision}, pages 21011--21021, 2023.

\bibitem[Li et~al.(2020{\natexlab{a}})Li, Bao, Zhang, Yang, Chen, Wen, and Guo]{li2020face}
Lingzhi Li, Jianmin Bao, Ting Zhang, Hao Yang, Dong Chen, Fang Wen, and Baining Guo.
\newblock Face x-ray for more general face forgery detection.
\newblock In \emph{Proceedings of the IEEE/CVF Conference on Computer Vision and Pattern Recognition}, 2020{\natexlab{a}}.

\bibitem[Li and Lyu(2018)]{li2018exposing}
Yuezun Li and Siwei Lyu.
\newblock Exposing deepfake videos by detecting face warping artifacts.
\newblock \emph{arXiv preprint arXiv:1811.00656}, 2018.

\bibitem[Li et~al.(2020{\natexlab{b}})Li, Yang, Sun, Qi, and Lyu]{li2019celeb}
Yuezun Li, Xin Yang, Pu Sun, Honggang Qi, and Siwei Lyu.
\newblock Celeb-df: A new dataset for deepfake forensics.
\newblock In \emph{Proceedings of the IEEE/CVF Conference on Computer Vision and Pattern Recognition}, 2020{\natexlab{b}}.

\bibitem[Liang et~al.(2022)Liang, Shi, and Deng]{liang2022exploring}
Jiahao Liang, Huafeng Shi, and Weihong Deng.
\newblock Exploring disentangled content information for face forgery detection.
\newblock In \emph{Proceedings of the European Conference on Computer Vision}, pages 128--145. Springer, 2022.

\bibitem[Liu et~al.(2021)Liu, Li, Zhou, Chen, He, Xue, Zhang, and Yu]{liu2021spatial}
Honggu Liu, Xiaodan Li, Wenbo Zhou, Yuefeng Chen, Yuan He, Hui Xue, Weiming Zhang, and Nenghai Yu.
\newblock Spatial-phase shallow learning: rethinking face forgery detection in frequency domain.
\newblock In \emph{Proceedings of the IEEE/CVF Conference on Computer Vision and Pattern Recognition}, 2021.

\bibitem[Luo et~al.(2023)Luo, Kong, Huang, Hu, Kang, and Kot]{luo2023beyond}
Anwei Luo, Chenqi Kong, Jiwu Huang, Yongjian Hu, Xiangui Kang, and Alex~C Kot.
\newblock Beyond the prior forgery knowledge: Mining critical clues for general face forgery detection.
\newblock \emph{IEEE Transactions on Information Forensics and Security}, 19:\penalty0 1168--1182, 2023.

\bibitem[Luo et~al.(2021)Luo, Zhang, Yan, and Liu]{luo2021generalizing}
Yuchen Luo, Yong Zhang, Junchi Yan, and Wei Liu.
\newblock Generalizing face forgery detection with high-frequency features.
\newblock In \emph{Proceedings of the IEEE/CVF Conference on Computer Vision and Pattern Recognition}, 2021.

\bibitem[Masi et~al.(2020)Masi, Killekar, Mascarenhas, Gurudatt, and AbdAlmageed]{masi2020two}
Iacopo Masi, Aditya Killekar, Royston~Marian Mascarenhas, Shenoy~Pratik Gurudatt, and Wael AbdAlmageed.
\newblock Two-branch recurrent network for isolating deepfakes in videos.
\newblock In \emph{Proceedings of the European Conference on Computer Vision}, 2020.

\bibitem[Nguyen et~al.(2019{\natexlab{a}})Nguyen, Fang, Yamagishi, and Echizen]{nguyen2019multi}
Huy~H Nguyen, Fuming Fang, Junichi Yamagishi, and Isao Echizen.
\newblock Multi-task learning for detecting and segmenting manipulated facial images and videos.
\newblock \emph{arXiv preprint arXiv:1906.06876}, 2019{\natexlab{a}}.

\bibitem[Nguyen et~al.(2019{\natexlab{b}})Nguyen, Yamagishi, and Echizen]{nguyen2019capsule}
Huy~H. Nguyen, Junichi Yamagishi, and Isao Echizen.
\newblock Capsule-forensics: Using capsule networks to detect forged images and videos.
\newblock In \emph{Proceedings of the IEEE International Conference on Acoustics, Speech, and Signal Processing}, 2019{\natexlab{b}}.

\bibitem[Ni et~al.(2022)Ni, Meng, Yu, Quan, Ren, and Zhao]{ni2022core}
Yunsheng Ni, Depu Meng, Changqian Yu, Chengbin Quan, Dongchun Ren, and Youjian Zhao.
\newblock Core: Consistent representation learning for face forgery detection.
\newblock In \emph{Proceedings of the IEEE/CVF Conference on Computer Vision and Pattern Recognition Workshop}, pages 12--21, 2022.

\bibitem[Qian et~al.(2020)Qian, Yin, Sheng, Chen, and Shao]{qian2020thinking}
Yuyang Qian, Guojun Yin, Lu Sheng, Zixuan Chen, and Jing Shao.
\newblock Thinking in frequency: Face forgery detection by mining frequency-aware clues.
\newblock In \emph{Proceedings of the European Conference on Computer Vision}, 2020.

\bibitem[Radford et~al.(2021)Radford, Kim, Hallacy, Ramesh, Goh, Agarwal, Sastry, Askell, Mishkin, Clark, et~al.]{radford2021learning}
Alec Radford, Jong~Wook Kim, Chris Hallacy, Aditya Ramesh, Gabriel Goh, Sandhini Agarwal, Girish Sastry, Amanda Askell, Pamela Mishkin, Jack Clark, et~al.
\newblock Learning transferable visual models from natural language supervision.
\newblock In \emph{International conference on machine learning}, pages 8748--8763. PMLR, 2021.

\bibitem[Rossler et~al.(2019)Rossler, Cozzolino, Verdoliva, Riess, Thies, and Nie{\ss}ner]{rossler2019faceforensics++}
Andreas Rossler, Davide Cozzolino, Luisa Verdoliva, Christian Riess, Justus Thies, and Matthias Nie{\ss}ner.
\newblock Faceforensics++: Learning to detect manipulated facial images.
\newblock In \emph{Proceedings of the IEEE/CVF Conference on International Conference on Computer Vision}, 2019.

\bibitem[Shiohara and Yamasaki(2022)]{shiohara2022detecting}
Kaede Shiohara and Toshihiko Yamasaki.
\newblock Detecting deepfakes with self-blended images.
\newblock In \emph{Proceedings of the IEEE/CVF Conference on Computer Vision and Pattern Recognition}, pages 18720--18729, 2022.

\bibitem[Sun et~al.(2022{\natexlab{a}})Sun, Liu, Yao, Sun, Chen, Ding, and Ji]{sun2022information}
Ke Sun, Hong Liu, Taiping Yao, Xiaoshuai Sun, Shen Chen, Shouhong Ding, and Rongrong Ji.
\newblock An information theoretic approach for attention-driven face forgery detection.
\newblock In \emph{European Conference on Computer Vision}, pages 111--127. Springer, 2022{\natexlab{a}}.

\bibitem[Sun et~al.(2022{\natexlab{b}})Sun, Yao, Chen, Ding, Li, and Ji]{sun2022dual}
Ke Sun, Taiping Yao, Shen Chen, Shouhong Ding, Jilin Li, and Rongrong Ji.
\newblock Dual contrastive learning for general face forgery detection.
\newblock In \emph{Proceedings of the AAAI Conference on Artificial Intelligence}, pages 2316--2324, 2022{\natexlab{b}}.

\bibitem[Sun et~al.(2023)Sun, Chen, Yao, Sun, Ding, and Ji]{sun2023towards}
Ke Sun, Shen Chen, Taiping Yao, Xiaoshuai Sun, Shouhong Ding, and Rongrong Ji.
\newblock Towards general visual-linguistic face forgery detection.
\newblock \emph{arXiv preprint arXiv:2307.16545}, 2023.

\bibitem[Tong et~al.(2022)Tong, Song, Wang, and Wang]{tong2022videomae}
Zhan Tong, Yibing Song, Jue Wang, and Limin Wang.
\newblock Videomae: Masked autoencoders are data-efficient learners for self-supervised video pre-training.
\newblock \emph{Advances in neural information processing systems}, 35:\penalty0 10078--10093, 2022.

\bibitem[Wang and Deng(2021)]{wang2021representative}
Chengrui Wang and Weihong Deng.
\newblock Representative forgery mining for fake face detection.
\newblock In \emph{Proceedings of the IEEE/CVF Conference on Computer Vision and Pattern Recognition}, 2021.

\bibitem[Wang et~al.(2022)Wang, Wu, Ouyang, Han, Chen, Jiang, and Li]{wang2022m2tr}
Junke Wang, Zuxuan Wu, Wenhao Ouyang, Xintong Han, Jingjing Chen, Yu-Gang Jiang, and Ser-Nam Li.
\newblock M2tr: Multi-modal multi-scale transformers for deepfake detection.
\newblock In \emph{Proceedings of the International Conference on Multimedia Retrieval}, pages 615--623, 2022.

\bibitem[Wang et~al.(2023)Wang, Bao, Zhou, Wang, and Li]{wang2023altfreezing}
Zhendong Wang, Jianmin Bao, Wengang Zhou, Weilun Wang, and Houqiang Li.
\newblock Altfreezing for more general video face forgery detection.
\newblock In \emph{Proceedings of the IEEE/CVF Conference on Computer Vision and Pattern Recognition}, pages 4129--4138, 2023.

\bibitem[Xu et~al.(2023)Xu, Liang, Jia, Yang, Zhang, and He]{xu2023tall}
Yuting Xu, Jian Liang, Gengyun Jia, Ziming Yang, Yanhao Zhang, and Ran He.
\newblock Tall: Thumbnail layout for deepfake video detection.
\newblock In \emph{Proceedings of the IEEE/CVF International Conference on Computer Vision}, pages 22658--22668, 2023.

\bibitem[Yan et~al.(2023{\natexlab{a}})Yan, Luo, Lyu, Liu, and Wu]{yan2023transcending}
Zhiyuan Yan, Yuhao Luo, Siwei Lyu, Qingshan Liu, and Baoyuan Wu.
\newblock Transcending forgery specificity with latent space augmentation for generalizable deepfake detection.
\newblock \emph{arXiv preprint arXiv:2311.11278}, 2023{\natexlab{a}}.

\bibitem[Yan et~al.(2023{\natexlab{b}})Yan, Zhang, Fan, and Wu]{yan2023ucf}
Zhiyuan Yan, Yong Zhang, Yanbo Fan, and Baoyuan Wu.
\newblock Ucf: Uncovering common features for generalizable deepfake detection.
\newblock \emph{arXiv preprint arXiv:2304.13949}, 2023{\natexlab{b}}.

\bibitem[Yan et~al.(2023{\natexlab{c}})Yan, Zhang, Yuan, Lyu, and Wu]{yan2023deepfakebench}
Zhiyuan Yan, Yong Zhang, Xinhang Yuan, Siwei Lyu, and Baoyuan Wu.
\newblock Deepfakebench: A comprehensive benchmark of deepfake detection.
\newblock \emph{arXiv preprint arXiv:2307.01426}, 2023{\natexlab{c}}.

\bibitem[Yan et~al.(2024)Yan, Yao, Chen, Zhao, Fu, Zhu, Luo, Yuan, Wang, Ding, et~al.]{yan2024df40}
Zhiyuan Yan, Taiping Yao, Shen Chen, Yandan Zhao, Xinghe Fu, Junwei Zhu, Donghao Luo, Li Yuan, Chengjie Wang, Shouhong Ding, et~al.
\newblock Df40: Toward next-generation deepfake detection.
\newblock \emph{arXiv preprint arXiv:2406.13495}, 2024.

\bibitem[Yang et~al.(2019)Yang, Li, and Lyu]{yang219exposing}
Xin Yang, Yuezun Li, and Siwei Lyu.
\newblock Exposing deep fakes using inconsistent head poses.
\newblock In \emph{Proceedings of the IEEE International Conference on Acoustics, Speech, and Signal Processing.}, 2019.

\bibitem[Zhang et~al.(2024)Zhang, Xiao, Li, Lin, Li, and Ge]{zhang2024learning}
Daichi Zhang, Zihao Xiao, Shikun Li, Fanzhao Lin, Jianmin Li, and Shiming Ge.
\newblock Learning natural consistency representation for face forgery video detection.
\newblock \emph{arXiv preprint arXiv:2407.10550}, 2024.

\bibitem[Zhao et~al.(2021)Zhao, Xu, Xu, Ding, Xiong, and Xia]{zhao2021learning}
Tianchen Zhao, Xiang Xu, Mingze Xu, Hui Ding, Yuanjun Xiong, and Wei Xia.
\newblock Learning self-consistency for deepfake detection.
\newblock In \emph{Proceedings of the IEEE/CVF Conference on International Conference on Computer Vision}, 2021.

\bibitem[Zheng et~al.(2021)Zheng, Bao, Chen, Zeng, and Wen]{zheng2021exploring}
Yinglin Zheng, Jianmin Bao, Dong Chen, Ming Zeng, and Fang Wen.
\newblock Exploring temporal coherence for more general video face forgery detection.
\newblock In \emph{Proceedings of the IEEE/CVF Conference on International Conference on Computer Vision}, pages 15044--15054, 2021.

\bibitem[Zhou et~al.(2017)Zhou, Han, Morariu, and Davis]{zhou2017two}
Peng Zhou, Xintong Han, Vlad~I. Morariu, and Larry~S. Davis.
\newblock Two-stream neural networks for tampered face detection.
\newblock In \emph{Proceedings of the IEEE/CVF Conference on Computer Vision and Pattern Recognition Workshop}, 2017.

\bibitem[Zhou et~al.(2021)Zhou, Wang, Liang, and Shen]{zhou2021face}
Tianfei Zhou, Wenguan Wang, Zhiyuan Liang, and Jianbing Shen.
\newblock Face forensics in the wild.
\newblock In \emph{Proceedings of the IEEE/CVF Conference on Computer Vision and Pattern Recognition}, pages 5778--5788, 2021.

\bibitem[Zhu et~al.(2021)Zhu, Wang, Fei, Lei, and Li]{zhu2021face}
Xiangyu Zhu, Hao Wang, Hongyan Fei, Zhen Lei, and Stan~Z Li.
\newblock Face forgery detection by 3d decomposition.
\newblock In \emph{Proceedings of the IEEE/CVF Conference on Computer Vision and Pattern Recognition}, pages 2929--2939, 2021.

\bibitem[Zhuang et~al.(2022)Zhuang, Chu, Tan, Liu, Yuan, Miao, Luo, and Yu]{zhuang2022uia}
Wanyi Zhuang, Qi Chu, Zhentao Tan, Qiankun Liu, Haojie Yuan, Changtao Miao, Zixiang Luo, and Nenghai Yu.
\newblock Uia-vit: Unsupervised inconsistency-aware method based on vision transformer for face forgery detection.
\newblock In \emph{Proceedings of European Conference on Computer Vision}, pages 391--407. Springer, 2022.

\bibitem[Zi et~al.(2020)Zi, Chang, Chen, Ma, and Jiang]{zi2020wilddeepfake}
Bojia Zi, Minghao Chang, Jingjing Chen, Xingjun Ma, and Yu-Gang Jiang.
\newblock Wilddeepfake: A challenging real-world dataset for deepfake detection.
\newblock In \emph{Proceedings of the 28th ACM international conference on multimedia}, pages 2382--2390, 2020.

\end{thebibliography}
}
\end{document}